# A Moonshot for AI Oracles in the Sciences

Bryan Kaiser[1*], Tailin Wu[2], Maike Sonnewald[3], Colin Thackray[4], and Skylar Callis[1]

[1] Los Alamos National Laboratory, X Computational Physics Division, Los Alamos, NM, USA
[2] Westlake University, Department of Engineering, Hangzhou, China
[3] University of California Davis, Department of Computer Science, Davis, CA, USA
[4] Harvard University, John A. Paulson School of Engineering and Applied Sciences, Cambridge, MA, USA

June 25, 2024

[*] corresponding author e-mail: bkaiser@lanl.gov

**Abstract**

Nobel laureate Philip Anderson and Elihu Abrahams once stated that, "even if machines did contribute to normal science, we see no mechanism by which they could create a Kuhnian revolution and thereby establish a new physical law." In this *Perspective*, we draw upon insights from the philosophies of science and artificial intelligence (AI) to propose necessary conditions of precisely such a mechanism for generating revolutionary mathematical theories. Recent advancements in AI suggest that satisfying the proposed necessary conditions by machines may be plausible; thus, our proposed necessary conditions also define a moonshot challenge. We also propose a heuristic definition of the intelligibility of mathematical theories to accelerate the development of machine theorists.

## 1 AI oracles are transforming the sciences

Recent progress in the field of artificial intelligence (AI), driven primarily by advances in deep learning,[1,2,3] will almost certainly continue to impact research practices in science.[4,5] In this *Perspective* we treat AI and deep learning as synonymous although deep learning is actually one sub-class of AI algorithms.[6,7] Arguably, the most salient impact of AI on scientific research is the emergence of a new type of scientific crisis characterized by the advent of accurate predictions that are *not* directly derived from highly intelligible theories, but rather by AI oracles.[5,8,9] Such oracles create perhaps the most acute manifestations of **anthropocentric predicament**,[10] in which humans struggle to comprehend the internal logic of an algorithm that enables it to surpass human abilities at certain tasks. Perhaps the most celebrated example of an AI oracle in science is AlphaFold 2, an algorithm that uses deep learning to predict protein-folding with superhuman accuracy but cannot directly produce theoretical contributions to the science of protein folding.[11,12,13,8] The list of AI oracles used to predict phenomena in scientific contexts includes examples in biology,[14,15] drug discovery,[16,14] oceanography,[17] climate science,[18,19] inorganic crystal discovery,[20] psychiatry,[21] prediction of material failure,[22] and continues to expand.

The epistemic problems arising from the widespread adoption of AI oracles in science engendered a new form of Kuhnian scientific crisis, where **Kuhnian crises** are periods of intense introspection, theoretical creativity, and disagreement that occur when discrepancies between empirical data and prevailing theories accumulate to a degree such that dissatisfaction with the prevailing theories becomes widespread.[23] In Kuhn's view, revolutions follow crises (e.g., past crises in physics include those resolved by the Newtonian, relativistic, and quantum revolutions), which are subsequently followed by periods of "normal science" in which the the epistemic consequences of prevailing theories of previous revolutions are explored to the greatest extent possible. We refer to the new form of crisis as the **oracular crisis** to underscore that the root of the crisis is revealed by comparisons of the abilities of AI oracles and scientists to achieve epistemic goals, rather than by the accumulation of discrepancies between prevailing theories and empirical data. The oracular crisis pushes formerly academic arguments regarding the relationship between scientific understanding, explanation, and prediction[24,25,26,27,9,28,10] out of philosophy and into scientific research practices.

Recent works of philosophy elucidate useful terminology for our discussion of the oracular crisis, the ensuing response, and the mechanism of Anderson-Abrahams.[29] **Oracles** are black box algorithms that conjure accurate predictions of empirical data,[25] where a **black box** is any device, computation, or process that transforms inputs into outputs without providing explicit or implicit references regarding the manner in which the inputs and outputs are linked.[26,24,6,7] Black boxes are epistemically opaque, where **epistemic transparency** is the degree[30] to which a device, computation, or process that transforms inputs into outputs can be represented as an intelligible theory.[31,10,27,28] Here, a **theory** is a predictive description of links between ontological entities (scientific ontologies will be discussed in greater detail below). A theory must contain one or more **empirical statements**[32] which are statements that relate two or more empirically measurable facts, e.g., measurable prediction $B$ (an output) can be quantified if given measurements of $A$ (an input). A theory $T$ is **intelligible** if it can be expressed in a form in which scientists "can recognize qualitatively characteristic consequences of $T$ without performing exact calculations."[25,33] **Models** are constructed from theories to explain and/or predict target systems,[24,25] and **algorithms** are finite, unequivocally-defined processes for calculating solutions to mathematical models and/or theories.[34,35,36]

The response to the oracular crisis has split into three epistemic perspectives, where each perspective is characterized by a different approach to addressing Sullivan's link uncertainty[27] within empirical statements. Sullivan proposed that the property of epistemic opacity arises from a dearth of epistemic links between oracle input and outputs rather than mathematical details and/or computational complexity. For example, factorials may be computed (or, "implemented") by either iterative or recursive algorithms that produce correct solutions.[27,37] The details of the implementation of the factorial algorithm lie outside of the epistemic perspective of scientific inquiries that require the calculation of factorials, where the **epistemic perspective** of a scientist is the assorted epistemic information (i.e., calculations, models, theories, prior knowledge) that a scientist uses to gain understanding of empirical target phenomena. Thus, a factorial calculator is an **implementation black box**[27] for scientific applications where its epistemic transparency/opacity lies outside of the scientist's epistemic perspective. Sullivan argues that the source of scientific discomfort with oracles arises from ambiguous epistemic links between input information and output predictions. **Link uncertainty** is uncertainty regarding the empirical validity of internal functions by which an oracle transforms inputs into predictions.[27] If one can a) *identify* the links (i.e., internal functions) and b) *empirically evaluate* some or all of the links of the oracle, then the oracle may become less opaque. In the next section we discuss how each of the three epistemic perspectives address link uncertainty.

## 2  Three epistemic perspectives emerge

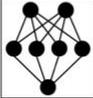

Figure 1: **Three epistemic perspectives have emerged in response to the oracular crisis.** In perspective 1, oracles are accepted "as is" as implementation black boxes. In perspective 2, the predictions of oracles and/or the internal states of oracles are constructed/examined to reduce link uncertainty by a) identifying and b) empirically evaluating internal epistemic links. The AI agent is viewed as (complex) empirical statement. In perspective 3, the internal states of an AI agent are accepted as implementation black boxes and the epistemic perspective shifts toward understanding target empirical data through the *output* of the AI agent, where the output is a relatively simple empirical statement with readily identifiable links. Perspective 3 also applies to human theorists.

The response to the oracular crisis is divided into three epistemic perspectives. The first perspective accepts the epistemic opacity of oracles and makes no effort to increase transparency. Hence, we call it the **oracular epistemic perspective**. Accurate predictions are valued over the understanding of the how the oracle internally links the target phenomena inputs and outputs (Figure 1). It is a "post-anthropocentric"[10] epistemic perspective[38] in which the AI agent is a predictive implementation black box.

The oracular perspective has many scientific applications as a tool for "fact finding." Fact finding is an essential component of exploratory science.[28,9,23] Scientists use oracles to detect and/or predict patterns that would otherwise be likely to remain hidden within large data sets.[15,20,39] Krenn *et al.*[5] refers to the ability of AI to find facts buried within large data sets as the use of AI as a "computational microscope," because microscopes reveal otherwise hidden information yet do not explain or provide epistemic links between the phenomena that they reveal. However, oracles reveal facts in the form of "phenomenon *A* is statistically associated with phenomenon *B*," which are susceptible to inductive bias[40,41] and therefore the epistemic links by which oracles predict that *A* arises from $B$[42,40] are highly uncertain because they are correlative. As Fersht[13] observed, despite the superhuman predictive accuracy of AlphaFold 2 at predicting protein structures,[11,12] "solving the pathway of protein folding, along with the dynamics of protein processes, is a different type of challenge from predicting protein structure."

The second perspective encompasses the diverse array of strategies that have been developed to increase understanding of the target phenomena by increasing the epistemic transparency of oracles through the reduction of link uncertainty. In this perspective, the AI agent is viewed as a complex empirical statement with high link uncertainty because the links by which the AI agent relates inputs and outputs are unvalidated (Figure 1). The link uncertainty is reduced by various methods that seek to identify and validate high-level links. The second perspective encompasses two approaches: interpretable artificial intelligence/machine learning (hereafter IAI)[43,44] and explainable artificial intelligence/machine learning (hereafter XAI).[45,46] We call this perspective the **XAI/IAI epistemic perspective**, noting that there is little consensus regarding mutually exclusive definitions of IAI and XAI.[47,48,49,43,50] IAI may be characterized as *ante hoc* strategies[51] for increasing the epistemic transparency of AI algorithms through the modification of algorithm design, the imposition of additional constraints, or other methods. Conversely, XAI may be characterized as the deployment of *post hoc* strategies[51,52,48,28] towards the same goal through the use of various forms of sensitivity analyses of feature salience relative to predictive accuracy, surrogate low-dimensional and/or linear models, or other methods. XAI and IAI are valuable tools for improving algorithm performance and informing algorithm selection for legal,[53,54] engineering,[55,56] and medical problems.[48,57] XAI and IAI also contribute to scientific advancements as components of novel research practices that integrate XAI and IAI models with theories and conventional computational models to identify epistemic links within empirical data and/or within model predictions.[5,38,19,58,19] XAI and IAI have increased in popularity since the maturation of deep learning in the 2010s (Figure 2), which suggests a possible rise in discomfort with the epistemic opacity of oracles.

The third perspective is the most nascent perspective and it defines the moonshot goal. The (human or AI) agent is regarded as an implementation black box, like the oracular perspective, but it produces output with readily identifiable epistemic links that are amenable to empirical evaluation: mathematical theories[59,60,61,62,63] (Figure 1). The output is a empirical statement or set of statements from which predictions may be calculated rather than a prediction itself. We refer to this perspective as the **Galilean epistemic perspective** because we restrict our discussion to mathematical theories for two reasons. First, it may be that all scientific theories can be expressed formally,[62,63] such as how Darwin's informal theory of evolution[64] was later expressed formally by Fisher.[65] Second, mathematical theories permit a special, prized form of intelligibility. We discuss the epistemic value of mathematical theories in the next section.

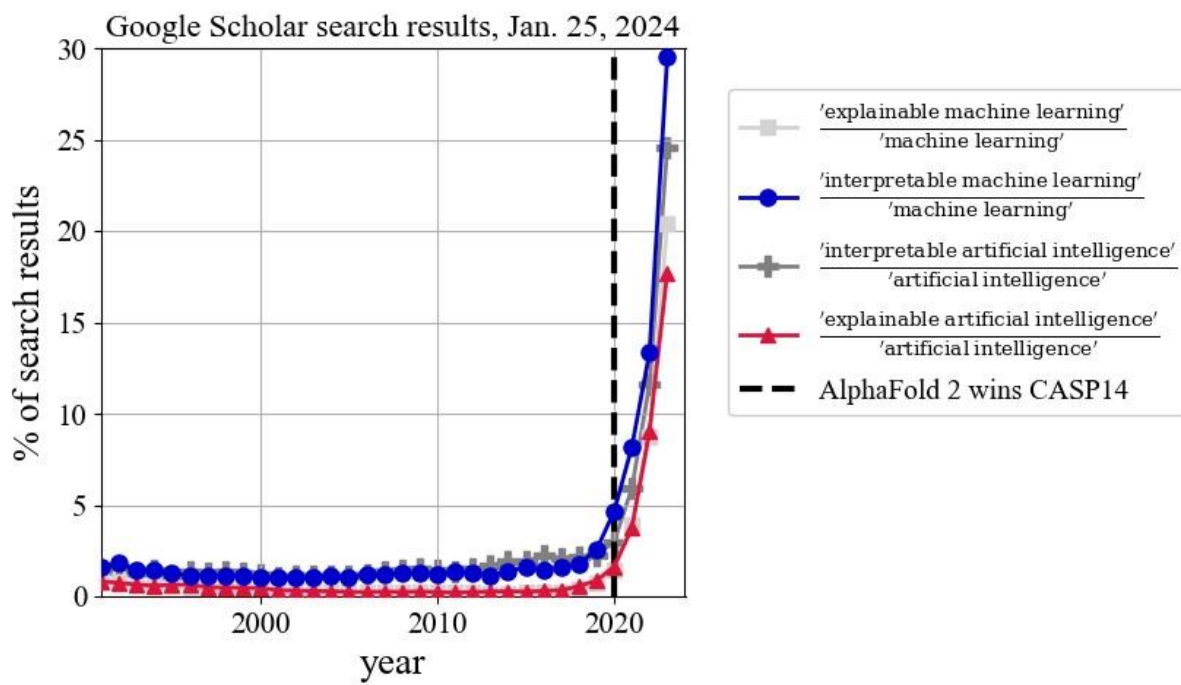

Figure 2: **Rising interest in the XAI/IAI epistemic perspective.** Percentages of Google Scholar search results recorded on January 25, 2024. 'interpretable machine learning' / 'machine learning' indicates the number of Scholar search results for "interpretable machine learning" divided by the number of Scholar search results for "machine learning," etc.

# 3 The Galilean epistemic perspective

Galileo was the first to champion the practice of formulating scientific theories in the language of formal mathematical systems.[66,67,68,69] There are at least three reasons why Galileo's preferred method for the formulation of theories became a pillar of modern science. First, mathematical theories enable empirical testing because they enable unambiguous and reproducible comparisons of theory-derived numerical predictions with empirical measurements.[32] Second, mathematics enable articulations of theories with high degrees of epistemic transparency. It is epistemic transparency, rather than objectivity, that enables modern scientific epistemology.[70] Subjectivity can never be completely purged from science,[40] regardless of the scientific commitment to remove personal and cultural biases from theories,[71,72] because theories necessarily include subjective and communal ontological concepts,[23,73,74] as we discuss below. Third, mathematical theories permit the representation of links between phenomena without invoking causes.[32,67] Like objectivity and absolute truth, causality is a problematic notion,[67,32] in part because it has no single definition. For example, both intuitive[75] (i.e., objects move bounded wholes that are incapable of "spooky interactions at a distance"[76]) and classical[77] (i.e., uniform space-time and a unidirectional arrow of time) definitions of causality conflict with modern physics theories.[78,79]

A **mathematical theory** is a predictive mathematical description of links between ontological entities. Thus, the intelligibility of a mathematical theory is dependent upon one's familiarity with the ontological concepts being linked together by mathematical statements and one's familiarity with mathematics. An **ontology** is a mutable list of concepts or things that exist.[7,73] For example, the scope of a personal ontology can be gauged by generating a list of all of the nouns that a person can name.[80] Ontological concepts link the symbols of theories to observable reality. Sellars proposed two useful ontologies: the **manifest image** is the ontology of all things germane to everyday life and the **scientific image** is an ontology defined by concepts from salient scientific theories.[80,81] The scientific image evolved as an offshoot of the manifest image.[80] Concepts in the scientific image are usually learned through scientific education. In AI, ontologies have been represented geometrically,[82] explicitly in symbolic form[7,83] and implicitly within large language models (LLMs) that enable high-fidelity mimicry of human concept manipulation through next token prediction.[84]

Ontologies are integral to all scientific theories in two ways. First, theories require the **ontological assumption**: observable reality can be split into separate entities that can be represented by symbols.[85] Crucially, the symbols that label different ontological concepts are shared.[83] The ontological assumption is tacitly invoked whenever different natural phenomena are distinguished from one another within a theory. Second, forms of the **regularity assumption** - that is, the assumption that a theory will apply as readily tomorrow/somewhere else as it does today/here - are so endemic to post-enlightenment scientific theories that it is generally left unstated. Regularity is a necessary assumption for theories because it is needed to dismiss Hume's problem of induction.[72,86,23,32,77] Hume argued that inductive reasoning justifies the regularity assumption, which itself justifies inductive reasoning, leading to circular logic. The assumption of regularity is invoked without empirical evidence (nor the hope of ever obtaining it), and it is the reason why theories can be proven false but never true.[32] The regularity assumption is tacitly invoked whenever predictions are derived from theories.

Mathematical theories are ontologically grounded if they include at least one ontologically grounded component, such as variables within an empirical statement. An **ontologically grounded** variable is a symbol in a mathematical theory that represents a concept or category, or a permutation thereof, from the scientific or manifest image. **Mathematical statements** are **ontologically neutral**, absent of grounding, unless otherwise specified. For example, it is impossible to include "$x$" within an epistemic perspective without grounding "$x$" to an ontological concept either explicitly or through context. If a theory designates variable $x$ as the number of trees per acre of forest, then $x$ is grounded within the theory. Mathematical operators, constants, and coefficients (e.g., $\pi$, coefficients of series expansions, etc.) are ontologically neutral, therefore, variables cannot inherit ontological grounding from mathematical operators, constants, or coefficients. Consider $\mathbf{x} = y \cdot \mathbf{z}$. The multiply and equal operators do not ontologically ground the variables. Therefore, $\mathbf{x} = y \cdot \mathbf{z}$ is not a mathematical scientific theory because we have not specified ontological concepts for variables $\mathbf{x}, y, \mathbf{z}$ and the mathematical operators are ontologically neutral.

Mathematical theories are empirical if they include at least one empirical statement. An **ontological statement** is a statement that relates one or more ontologically grounded variables. *All empirical statements are ontological but not all ontological statements are empirical* (e.g., the regularity assumption discussed previously). A variable is empirical if it is ontologically grounded to a concept that is empirically measurable. An **empirical fact**, hereafter fact, is an instance (e.g., scalars) or a collection of instances (e.g., tensors) of measurements that correspond to ontologically grounded variable(s). At minimum, a theory must contain an empirical statement that relates two empirical facts to expose the theory to empirical tests.[32] The relationships between empirical statements, ontological statements, and mathematical statements are shown in Figure 3.

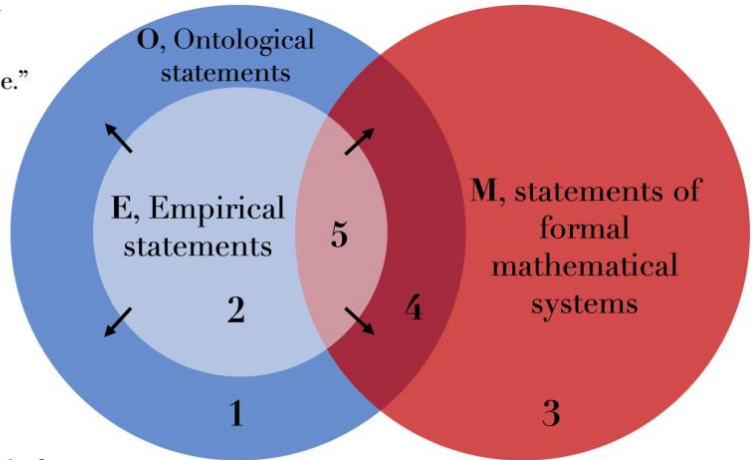

1) O − (E ∪ M): Non-empirical ontological statements, "our universe is part of a multiverse."
2) E − M: Quantitative statements/facts, "Earth's diameter is ~ 13,000 km."
3) M − O: Pure mathematical statements, "x = y²."
4) M ∩ O − E: Non-empirical mathematical & ontological statements, e.g., the Drake equation.
5) M ∩ E: Empirical mathematical statements, e.g., Newton's laws.

Figure 3: **Theoretical statements.** Purely ontological statements (region 1) are not presently measurable. All empirical statements (region 2) are a subset of ontological statements because an entity must be identified/labeled to be measurable. Purely mathematical statements (region 3) are non-empirical abstractions. Ontological mathematical statements, such as the Drake equation (a prediction of the number of extraterrestial civilizations in the Milky Way Galaxy that presently cannot be measured), reside in region 4. Galilean scientific theories reside in region 5; empirical mathematical statements that relate measurable quantities. The black arrows represent how new technologies enable new measurements. E.g., the formerly (unmeasureable) ontological statements of gravitational waves, derived from the theory of general relativity, became empirical statements of general relativity with the advent of technologies capable of measuring gravitational waves.[87,88]

Ontologies are defined *externally* to the agents that manipulate and express concepts in the form of ontologically grounded symbols.[23,74] Thus, the external grounding of ontologies eliminates the necessity of internal concept definition.[89] A theorizing agent need not disclose their internal encoding of ontologies and epistemic links to propose intelligible theories, just as scientists need not submit measurements of their brain activity when submitting a theory for publication. Any agent (human or AI) capable of representing relationships between ontological concepts with mathematics can produce a theory that is intelligible to other agents with whom it shares a set of ontological concepts, or, "transfer its understanding,"[5] regardless of the internal representations of ontological concepts within the agents.

New ontological concepts are only accepted if they reference or modify pre-existing concepts. For example, if we invent the concept of a "barbarbar" and the reader asks what it is, we will be forced to *ground* the concept of the "barbarbar" to an ontology already known to the reader with a statement such as, "It's a mammal." Another example is Einstein's proposal of the new (in 1914) ontological concept of space-time, a concept directly derived from the combination of differential geometry of the theory together with the pre-existing concepts of fixed space and time. In this manner, new scientific concepts can be formed through novel combinations of pre-existing concepts.[90]

The mathematical operators, constants, and coefficients can *modify* the previous definitions of the relevant ontological concepts. Consider the variables and operators in Newton's three laws of motion. Newton modified the pre-existing ontological concepts of force, mass, momentum, and acceleration (drawn from the scientific image of his era[91,92]) by proposing specific mathematical relationships between them. The multiplication and equality operators in Newton's second law of motion, **F** = $m \cdot$ **a**, defined new (in 1687) mechanistic relationships between the variables. Newton's version of the concepts of force, mass, and acceleration was created through the proposal of a novel mathematical relationship between pre-existing ontological concepts.

We propose that the modification of ontologically grounded concepts by new mathematical theories is the process by which the Galilean branch of the scientific image expands. We call this process **Wignerian expansion** because Wigner previously articulated this idea: if a theory accurately predicts facts, then a scientist cannot help but to "jump at the conclusion that the connection *is* that discussed in mathematics."[93] Wignerian expansion is the product of novel concept associations[90] and/or novel mathematical relationships between concepts. The Galilean branch of the scientific image can also contract as some ontological concepts are discarded when the theories they are a part of are discarded, e.g., luminiferous aether.[23]

In contrast to fitted models (i.e., symbolic regression) in which ontological concepts are statistically defined, revolutionary scientific theories can be - and often are - ontologically extrapolative. **Ontological extrapolation** is the successful prediction of the existence of phenomena that is not measureable at the time the theory is generated. For mathematical theories, it is often a consequence of Wignerian expansion because that is the process by which new ontological phenomena are conceptualized. Examples include a) predictions of phases of Venus by Copernicus' theory of celestial bodies that were validated sixty years after his death,[23] b) predictions of a white spot at the center of a shadow of a circular disk by Poisson's *reductio ad absurdum* formulation of Fresnel's wave theory to the surprise of Fresnel and his contemporaries,[23] c) the prediction of gravitational waves by Einstein's theory of general relativity nearly 100 years before the waves were observed,[88,87] and d) the prediction of the existence of neutrinos and the Higgs boson approximately 25 and 50 years before they were observed, respectively.[94]

# 4 Necessary conditions of the Anderson-Abraham mechanism

Anderson and Abraham's skepticism of the possibility of machine theorization[29] was justified in 2009 because no contemporary AI agent demonstrated a modicum of capability to manipulate ontological concepts or formally reason in a flexible and human-like manner, thus, there was no foreseeable "mechanism" by which a revolutionary mathematical theory could be generated by a machine. Such a mechanism no longer appears to be so far fetched. LLMs have recently demonstrated a modicum of ability to informally (i.e., through natural language)[84,95,96] and formally reason,[97,98,99,100,101,22,102] or at least, appear to exhibit behavior that is functionally equivalent to limited forms of reasoning.[96] We hypothesize that informal and formal reasoning capabilities of AI emerge as the number of learned parameters and computational resources increases[103,104,105,106] as an epiphenomenological byproduct of sufficiently complex forms of information processing. This hypothesis is merely a version of Turing's conjecture that a sufficiently complex information processing system may, when prompted, "give rise to a whole 'theory' consisting of secondary, tertiary and more remote ideas."[107] However, there are many reasons to suspect that the emergence of reasoning in deep learning may be illusory;[108,109,110] further research is necessary to determine the validity of the hypothesis and a complete review of the matter is beyond the scope of this *Perspective*.

Assuming that the ability of AI to manipulate ontological concepts and formally reason will develop in the years ahead, we propose three necessary conditions for an AI to be capable of generating revolutionary mathematical theories:

1. The ability to formally conjecture, derive, and prove mathematical statements;

2. The ability to represent, combine, and alter ontological concepts from the scientific and manifest images;

3. The ability to combine 1 and 2 to form empirical statements by manipulating mathematics to be consistent with ontological concepts and vice versa.

The necessary conditions distinguish symbolic regression,[60,111] in which ontological concepts are prescribed and therefore necessary conditions 2 and 3 remain unsatisfied, from the process of generating a revolutionary mathematical theory. An agent that satisfies the necessary conditions may also satisfy Krenn *et al.*'s second sufficiency condition for scientific understanding, "an AI gained scientific understanding if it can transfer its understanding to a human expert,"[5] if the generated theories are highly intelligible. We discuss the intelligibility of mathematical theories in the next section.

# 5 Galilean intelligibility

Scientific theories are evaluated by multiple criteria which vary in salience across scientific communities, individuals, and time. Empirical criteria include the scope of applicability[32,72] and predictive accuracy of a theory within the scope of phenomena to which it applies.[71,72,112] Non-empirical criteria include aesthetic concerns[94] such as naturalness,[113,114] beauty[115,116,117] and simplicity.[118] A comprehensive discussion of these criteria is beyond the scope of this *Perspective*. Instead, we present a definition of the intelligibility of mathematical theories because intelligible mathematical theories are central to the Galilean epistemic perspective.

A theory is **Galilean intelligible** if it is a set of mathematical statements that contain at least one empirical statement. Galilean intelligibility is a measure of the epistemic transparency of mathematical theories, and it is a special case of de Regt's criterion for the intelligiblity of theories[25,33] (Section 1). The degree of Galilean intelligibility $I$ of an empirical statement of a theory is determined by the ratio of the number of empirical constants to the number of ontologically grounded variables, hereafter variables, and is formulated as

$$I = 1 - \frac{N_E}{N_O} \in (-\infty, 1], \tag{1}$$

where $N_E$, the number of empirical constants, and $N_O$, the number of variables, are both natural numbers. A theory with maximum intelligibility has no empirical constants ($N_E = 0, I = 1$). $N_O \geq 2$ for empirical statements (see Sections 1 and 3), whereas $N_E \geq 0$. Increasing $I$ corresponds to increasing intelligibility, i.e., increasing epistemic transparency, such that perfect intelligibility takes the value of $I = 1$. When the ratio of empirical constants to variables is high/low, empirical statements have low/high epistemic transparency, such that empirical statements characterized by $I \ll 0$ become epistemically opaque. I.e., $I \ll 0$ statements are reduced to epistemically vague interpretations like "phenomenon $A$ is statistically associated with phenomenon $B$." This is the reason that scientists prefer to minimize the number of empirical constants relative to variables and ultimately hope to create new theories with no empirical constants.[116,32,94] Mathematical sophistication, computational complexity, empirical validity, scope of applicability, etc., are all separate criteria for evaluating empirical statements from Galilean intelligibility.

**Empirical constants** are defined as variables that are assigned a constant value in an empirical statement through targeted observational estimates (e.g., physical constants like Planck's, the speed of light, etc.) and/or through regression over empirical data (e.g., the observed transistor doubling period of Moore's law[119]). Empirical constants may include the weight, bias, and any other fitted tensors of deep neural networks trained on empirical data. Neither the empirical robustness (i.e., the scope of phenomena to which it applies) nor the empirical validity (i.e., the predictive accuracy of the empirical statement in which it appears) affect an empirical constant's status as an empirical constant. Mathematical constants, such as $\pi$ and $e$, are not empirical constants because they do not require empirical measurements to determine their values. All constants that can be algebraically combined must be combined when determining the number of empirical constants $N_E$ in an empirical statement.

To estimate $N_O$ and $N_E$, variables and empirical constants must be counted from the dyadic and/or continuous form of the empirical statement. The dimensions of variables and empirical constants do *not* contribute to $N_O$ or $N_E$, respectively, *nor* do the number of discrete samples of the variables and empirical constants. For example, if an empirical constant appears in an empirical statement that may be used to predict the dynamics of Avogadro's number of molecules, it only counts as one empirical constant even thought it may be repeated in $6 \cdot 10^{23}$ calculations. Similarly, a three-dimensional velocity counts as one variable.

Distinguishing empirical constants from variables can depend on the application of the empirical statement. For example, mass may be a constant (e.g., predicting the motion of a pendulum) or a variable (e.g., predicting mass-energy conversion with $E = mc^2$) in different contexts. The exact value of $I$ is always approximate and context dependent. Equation 1 is valuable for the estimation of the epistemic transparency of one empirical statement *relative* to another empirical statement. Comparisons require consistent application of ontological concepts. Two empirical statements have significantly different transparency when the difference in respective $I$ values are robust to ±1 variations in the counts $N_E$ and $N_O$. Therefore, all of the empirical statements in Figure 4 with $I > -1$ (Equations 6-19) may be viewed as equally highly intelligible. The empirical statements and estimates of $N_O$, $N_E$, and $I$ for each empirical statement in Figure 4 are shown in Table 1.

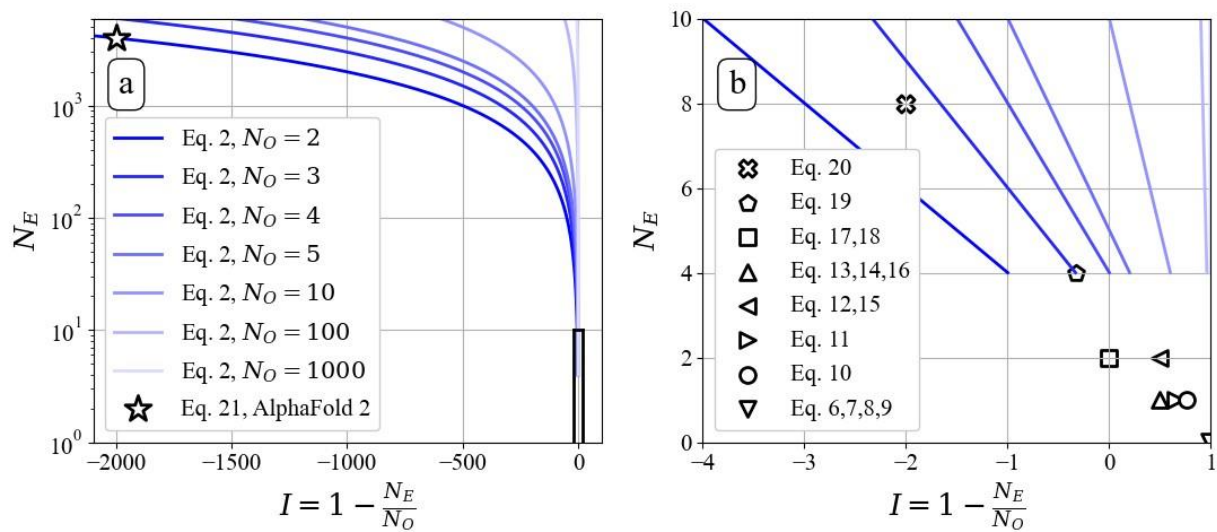

Figure 4: **The Galilean intelligibility of empirical statements.** *a* depicts the Galilean intelligibility of simple fully connected neural networks with different numbers of ontological variables $N_O$ for the network described by Equations 2 and 3. The Galilean intelligibility of AlphaFold 2 is represented by the star. It has approximately $N_h$ = 2000 hidden layers (personal communication, John Jumper) and links amino acid sequences to protein structures, therefore we estimate $N_E \approx 4000$, $N_O = 2$, $I \approx -2000$. Figure *b* corresponds to the zoom box drawn on Figure *a* near $I$ = 0. The empirical statements (Equations 2, 6-21) are described in Table 1 and the supplemental information.

The Galilean intelligibility of a conventional, fully connected, feed-forward, trained artificial neural network is estimated as follows. A simple fully connected artificial neural network can be expressed as[120]

$$\mathbf{y} = \beta_{N_h} + \Omega_{N_h} f_a(\beta_{N_h-1} + \Omega_{N_h-1} f_a(...\beta_2 + \Omega_2 f_a(\beta_1 + \Omega_1 f_a(\beta_0 + \Omega_0 \mathbf{x}))...)),, \quad (2)$$

$$= f_\theta(\mathbf{x}),$$

where $\mathbf{\Omega}_i$ are weight tensors, $\mathbf{\beta}_i$ are bias tensors, $f_a()$ are activation functions, $f_\theta()$ represents the entire network, and $N_h$ is the number of hidden layers. $N_E$ is determined by the number of hidden layers because the weight and bias tensors of the hidden layers cannot be combined. For Equation 2 the number of empirical constants is

$$N_E = 2(N_h + 1) \geq 4 \qquad \forall N_h \geq 1, \tag{3}$$

where there is one weight and bias tensor per hidden layer plus one weight and bias tensor for the output layer. Therefore, the Galilean intelligibility of conventional deep neural networks described by Equation 2 is given by

$$I_\theta = 1 - \frac{2(N_h + 1)}{N_O}. \tag{4}$$

If $N_O = 2$, the network is an empirical statement that relates two variables, such as an amino acid sequence ($\mathbf{x} = \mathbf{a}$) and subsequent protein structure ($\mathbf{y} = \mathbf{p}$) in the case of AlphaFold, then $I = -N_h$. Actual neural networks such as AlphaFold 2 possess much more complex architectures than the one in Equation 2. Nevertheless, the number of hidden layers/tensors $N_h$ is a reasonably conservative proxy for the number of empirical constants in a deep neural network. Many neural networks relate small numbers of ontological variables such that $N_h \gg N_O$, therefore

$$I_\theta = -O(N_h), \tag{5}$$

is an approximation of the Galilean intelligibility of trained neural networks that satisfy $N_h \gg N_O$.

Equation 4 and Figure 4 suggest that neural networks can be highly Galilean intelligible ($-1 \leq I \leq 1$) if they have only few hidden layers ($N_h = O(1)$) and/or relate many variables ($N_O \gg 2$). Otherwise, neural networks have low Galilean intelligibility. Increasing the number of variables merely for the sake of adding variables is problematic because extraneous variables that are not required for the network to learn epistemic links may create training/performance issues, such as requiring the addition of extraneous training data. Oracles such as AlphaFold 2 contain $O(10^3)$ hidden layers (personal communication, John Jumper); thus, the task of choosing $O(10^3)$ additional variables that are salient to protein folding prediction may be insurmountable. In addition, predictive performance may require networks to be "overparameterized,"[120] where the number of network parameters is $O(10^2)$ times the number of training samples. Overparameterization implies that networks with small numbers of hidden layers $N_h = O(1)$ will sacrifice predictive accuracy for Galilean intelligibility in many cases.

Table 1: Empirical statements from high profile theories, plotted in Figure 4. ([*] indicates relevant equation details provided in supplemental information.)

| Eq. | Empirical statement | Variables | Empirical const. | $I$ |
| 6 | Newton's first law | 1: force **F** | 0 | 1 |
| | $\mathbf{F} = 0$ (6) | | | |
| 7 | Newton's second law | 3: force **F**, mass $m$, acceleration **a** | 0 | 1 |
| | $\mathbf{F} = m\mathbf{a}$ (7) | | | |
| 8 | Newton's third law | 3: time $t$, momentums of two distinct objects, $\mathbf{p}_1, \mathbf{p}_2$ | 0 | 1 |
| | $\frac{d\mathbf{p}_1}{dt} = \frac{d\mathbf{p}_2}{dt}$ (8) | | | |
| 9 | compressible Navier-Stokes, variable viscosity[121] | 7: velocity **u**, position **x**, pressure $p$, time $t$, viscosity $\mu$, density $\rho$ | 0 | 1 |
| | $\frac{\partial(\rho\mathbf{u})}{\partial t} + \nabla \cdot (\rho\mathbf{u} \otimes \mathbf{u}) = -\nabla p + \nabla \cdot \left(\mu\left(\nabla\mathbf{u} + (\nabla\mathbf{u})^T - \frac{2}{3}(\nabla \cdot \mathbf{u})\mathbf{I}\right)\right)$ (9) | | | |
| 10 | Newton's law of gravitation | 4: force **F**, mass of two distinct objects $M, m$, distance between the objects $r$ | 1: gravitational constant $G$ | 3/4 |
| | $\mathbf{F} = \frac{GMm}{r^2}\mathbf{r}$ (10) | | | |
| 11 | Nonrelativistic, no spin Shrö̈dinger equation[122] | 3: wave function $\phi$, particle energy Hamiltonian H, time $t$ | 1: reduced Planck constant $\hbar$ | 2/3 |
| | $i\hbar\frac{\partial \phi}{\partial t} = H\phi$, (11) | | | |
| 12 | incompressible Navier-Stokes[121] | 4: velocity **u**, position **x**, pressure $p$, time $t$ | 2: density $\rho$, viscosity $\nu$ | 1/2 |
| | $\frac{\partial \mathbf{u}}{\partial t} + \mathbf{u} \cdot \nabla\mathbf{u} = -\frac{1}{\rho}\nabla p + \nu\nabla^2\mathbf{u}$ (12) | | | |
| 13 | Zipf's law[123] | 2: word frequency $\omega$, word rank $r$ | 1: power of rank $a$ | 1/2 |
| | $\omega = r^a$ (13) | | | |
| 14 | Moore's law[119]* | 2: number of transistors $N$, years $Y$ | 1: doubling period of transistors $P$ | 1/2 |
| | $N = 2^{Y/P}$ (14) | | | |
| 15 | incompressible Reynolds stress transport equation[124]* | 4: velocity **u**, position **x**, pressure $p$, time $t$ | 2: density $\rho$, viscosity $\nu$ | 1/2 |

$$\frac{\partial \tau_{ij}}{\partial t} + U_k \frac{\partial \tau_{ij}}{\partial x_k} = -\tau_{ik}\frac{\partial U_j}{\partial x_k} - \tau_{jk}\frac{\partial U_i}{\partial x_k} + 2\nu \frac{\partial u'_i}{\partial x_k}\frac{\partial u'_j}{\partial x_k} - \Pi_{ij} + \frac{\partial}{\partial x_k}\left(\nu \frac{\partial \tau_{ij}}{\partial x_k} + C_{ijk}\right) \quad (15)$$

| | | | | |
|---|---|---|---|---|
| 16 | Einstein's mass-energy equivalence | 2: energy $E$, mass $m$ | 1: speed of light $c$ | 1/2 |
| | $E = mc^2$ (16) | | | |
| 17 | Zipf-Mandelbrot law[123] | 2: word frequency $\omega$, word rank $r$ | 2: shift of rank $b$, power of rank $a$ | 0 |
| | $\omega = (r + b)^a$ (17) | | | |
| 18 | radiative & collisional nonlocal thermodynamic equilibrium rate equation[125] | 2: number of particles fluxing in/out $n_{\text{in/out}i}$ | 2: transition probabilities of in/out particles $P_{ij\text{in/out}}$ | 0 |
| | $n_{\text{in}i}P_{ij\text{in}} = n_{\text{out}i}P_{ij\text{out}}$ (18) | | | |
| 19 | standard $k$-$\epsilon$ turbulence model, $\epsilon$ equation[124*] | 3: velocity $\mathbf{u}$, position $\mathbf{x}$, time $t$ | 4: viscosity $\nu$, tunable coefficients ($C_{\epsilon1}, C_{\epsilon2}, C_\mu/\sigma_\epsilon$) | -1/3 |
| | $\frac{\partial \epsilon}{\partial t} + U_j \frac{\partial \epsilon}{\partial x_j} = C_{\epsilon1}\frac{\epsilon}{k}\tau_{ij}\frac{\partial U_i}{\partial x_j} - C_{\epsilon2}\frac{\epsilon^2}{k} + \frac{\partial}{\partial x_j}\left(\nu + \frac{C_\mu}{\sigma_\epsilon}\frac{k^2}{\epsilon}\frac{\partial k}{\partial x_j}\right)$ (19) | | | |
| 20 | Spalart-Allmaras turbulence model[124*] | 4: velocity $\mathbf{u}$, position $\mathbf{x}$, time $t$, distance from surface $d$ | 8: viscosity $\nu$, coefficients $c_{b1}, c_{b2}, c_{v1}, c_{w2}, c_{w3}, \kappa, \sigma$ | -2 |
| | $\frac{\partial \tilde{\nu}}{\partial t} + U_j \frac{\partial \tilde{\nu}}{\partial x_j} = c_{b1}\tilde{S}\tilde{\nu} - c_{w1}f_w\left(\frac{\tilde{\nu}}{d}\right)^2 + \frac{1}{\sigma}\frac{\partial}{\partial x_k}\left((\nu + \tilde{\nu})\frac{\partial \tilde{\nu}}{\partial x_k}\right) + \frac{c_{b2}}{\sigma}\frac{\partial \tilde{\nu}}{\partial x_k}\frac{\partial \tilde{\nu}}{\partial x_k}$ (20) | | | |
| 2 | fully connected, trained, artificial neural network relating an amino acid sequence to a protein structure | 2: $\mathbf{x},\mathbf{y}$ each corresponds to 1 ontological variable | $N_E = 2(N_h + 1)$: each hidden layer has 1 weight $\mathbf{\Omega}$ and 1 bias $\boldsymbol{\beta}$ tensor | $-N_h$ |
| 21 | AlphaFold 1[126] | 2: amino acid sequence $\mathbf{a}$, protein structure $\mathbf{p}$ | $N_E \approx 1322$: ($N_h \approx$ 220 blocks · 3 layers per block = 660) | $-O(10^3)$ |
| | $\mathbf{p} = f_\theta(\mathbf{a})$ (21) | | | |
| 21 | AlphaFold 2[11] | 2: amino acid sequence $\mathbf{a}$, protein structure $\mathbf{p}$ | $N_E \approx 2000$ | $-O(10^3)$ |

# 6 Achieving the moonshot

The ultimate manifestation of the moonshot goal is the generation of prized mathematical theories, such as general relativity, by machines. However, machine theorization (Sections 1, 2, and 3) is perhaps more valuable in mundane applications. Just as human scientists may contribute to theoretical advancements through the process that Kuhn referred as to the "mop up operations" of normal science,[23] in which further consequences of prevailing theories are derived, empirically tested, etc., so may machine theorists contribute to normal science in myriad ways.

We proposed three necessary conditions that an AI agent must satisfy to achieve the moonshot (Section [4]). If an AI agent achieves the moonshot, then the theories it produces will be Galilean intelligible (Section [5]), a special case of de Regt's criterion for intelligibility.[25,33] However, we have argued that low Galilean intelligibility is a manifestation of low epistemic transparency, that AI oracles function as low Galilean intelligibility theories, and that scientists value theories with high Galilean intelligibility. Accordingly, we defined Galilean intelligibility as a quantitative heuristic with the intention that it may be applied either explicitly in algorithm design/training or implicitly through methods such as learning with human feedback.[127] Looking forward, we envision a future in which the oracular, XAI/IAI, and Galilean epistemic perspectives each contribute to advancements in science.

# 7 Acknowledgements

This work was performed under the auspices of U.S. Department of Energy (DOE). Financial support comes partly from Los Alamos National Laboratory (LANL), the DOE's Advanced Simulation and Computing program and the Office of Experimental Science's Advanced Diagnostics program. This work is approved for unlimited release, LA-UR-23-31369. LANL, an affirmative action/equal opportunity employer, is managed by Triad National Security, LLC, for the National Nuclear Security Administration of the U.S. Department of Energy under contract 89233218CNA000001. The first author acknowledges Ismael Boureima (LANL), Kyle Hickmann (LANL), Elisabeth Moore (Pacific Northwest National Laboratory), Matthew Osman (University of Cambridge), Rosalyn Rael (LANL), Juan Saenz (LANL), Greg Salvesen (LANL), and Michael Strevens (New York University) for their contributions to the ideas discussed in this manuscript.

# 8  Supplemental information

## 8.1  Turbulence models

The turbulence models (Equations 15, 20, and 19) are formulated using the following relations, where the overbar represents averaging:

$$\mathbf{u} = \mathbf{U} + \mathbf{u}',$$
$$\overline{\mathbf{u}} = \mathbf{U},$$
$$\tau_{ij} = -\overline{u'_i u'_j},$$
$$p = P + p',$$
$$\overline{p} = P,$$
$$\nu_T = \tilde{\nu} f_{v1},$$
$$f_{v1} = \frac{\chi^3}{\chi^3 + c_{v1}^3},$$
$$f_{v2} = 1 - \frac{\chi}{1 + \chi f_{v1}},$$
$$f_w = g\left(\frac{1 + c_{w3}^6}{g^6 + c_{w3}^6}\right)^{1/6},$$
$$\chi = \frac{\tilde{\nu}}{\nu},$$
$$\tilde{S} = S + \frac{\tilde{\nu}}{\kappa^2 d^2} f_{v2},$$
$$S = \sqrt{2\Omega_{ij}\Omega_{ij}},$$
$$\Omega_{ij} = \frac{1}{2}\left(\frac{\partial U_i}{\partial x_j} - \frac{\partial U_j}{\partial x_i}\right),$$
$$r = \frac{\tilde{\nu}}{\tilde{S}\kappa^2 d^2},$$
$$g = r + c_{w2}(r^6 - r),$$
$$c_{w1} = \frac{c_{b1}}{\kappa} + \frac{1 + c_{b2}}{\sigma},$$
$$k = \frac{1}{2}\overline{u'_i u'_i},$$
$$\epsilon = 2\nu \overline{\frac{\partial u'_i}{\partial x_k}\frac{\partial u'_i}{\partial x_k}},$$
$$\Pi_{ij} = \overline{\frac{p'}{\rho}\left(\frac{\partial u'_i}{\partial x_j} + \frac{\partial u'_j}{\partial x_i}\right)},$$
$$C_{ijk} = \overline{u'_i u'_j u'_k} + \frac{\overline{p' u'_i}}{\rho}\delta_{jk} + \frac{\overline{p' u'_j}}{\rho}\delta_{ik}.$$

## 8.2 Moore's law

Moore's law (Equation 14) requires

$$N = n/n_0,$$
$$Y = y - y_0,$$

where $n$ is the number of transistors for year $y$ and $n_0$ is the number of transistors in the first year $y_0$. Therefore, $n \geq n_0$ and $y \geq y_0$.